\documentclass[11pt]{article}
\usepackage[T1]{fontenc}
\usepackage[utf8]{inputenc}
\usepackage{arxiv_card}
\usepackage{amsmath,amssymb,amsthm,booktabs,tabularx,array,graphicx}
\usepackage{subcaption}
\usepackage{algorithm}
\usepackage{algorithmic}
\usepackage{pifont}
\usepackage{hyperref}
\hypersetup{colorlinks=true,linkcolor=cardaccent,citecolor=cardaccent,urlcolor=cardaccent}
\usepackage{multirow}
\usepackage{url}
\usepackage{enumitem}
\newtheorem{myrq}{\textbf{RQ}}
\newtheorem{remark}{Remark}
\newcolumntype{L}[1]{>{\raggedright\arraybackslash}p{#1}}
\newcolumntype{Y}{>{\raggedright\arraybackslash}X}

\title{Bayes-Adaptive MDPs for Rollout Allocation to Enhance GRPO Training}
\runningtitle{Bayes-Adaptive MDPs for Rollout Allocation in GRPO}
\author{
Faeze Moradi Kalarde$^{1, 2}$, Armin Toroghi$^{1, 2}$, Nam Ho Koh$^{3}$, Doo Hwan Hwang$^{3}$, Seungjin Kim$^{3}$, Kee-Eung Kim$^{3}$, Scott Sanner$^{1, 2}$
}
\hypersetup{
  pdftitle={Bayes-Adaptive MDPs for Rollout Allocation to Enhance GRPO Training},
  pdfauthor={Faeze Moradi Kalarde, Armin Toroghi, Nam Ho Koh, Doo Hwan Hwang, Seungjin Kim, Kee-Eung Kim, Scott Sanner}
}
\affiliation{$^{1}$University of Toronto \\ $^{2}$Vector Institute of AI \\ $^{3}$Korea Advanced Institute of Science $\&$ Technology}
\date{}

\begin{document}
\begin{abstract}
Group Relative Policy Optimization (GRPO) has demonstrated strong performance in LLM post-training. However, it allocates a uniform rollout budget to every training question, ignoring variations in question difficulty. As a result, highly difficult questions may receive too few rollouts to generate informative positive reward signals, limiting their contribution to training, while moderately difficult questions often receive more rollouts than necessary. 
In this work, we formulate the problem of difficulty-aware rollout budget allocation to maximize training utility under the same total rollout budget as standard GRPO. We propose AdaRoll-GRPO, which reformulates rollout allocation as a sequential decision-making problem and leverages a Bayes-adaptive Markov decision process (MDP) to learn an adaptive allocation policy. The learned policy is computed offline and applied during GRPO training without introducing additional online optimization overhead. Extensive experiments on multiple reasoning datasets and LLM architectures demonstrate that AdaRoll-GRPO consistently improves GRPO training performance under a fixed rollout budget, highlighting the effectiveness of adaptive rollout allocation for GRPO in LLM post-training.


\end{abstract}

\maketitle
\section{Introduction}

Reinforcement Learning (RL)-based post-training has emerged as a key paradigm for enhancing the capabilities of Large Language Models (LLMs) by optimizing them through task-specific feedback signals. Among RL methods for LLM post-training, Group Relative Policy Optimization (GRPO)~\citep{shao2024deepseekmath, guo2025deepseek} has recently gained significant attention due to its computationally efficient advantage estimation. Unlike traditional policy optimization methods such as Proximal Policy Optimization (PPO)~\citep{schulman2017proximal}, which require training and maintaining a separate value model to estimate advantages, GRPO estimates advantages directly from the relative rewards of multiple responses sampled for the same prompt. By eliminating the value model, GRPO significantly reduces the computational and memory overhead of RL post-training.
\begin{figure}[h]
    \centering
    \includegraphics[width=0.9\textwidth]{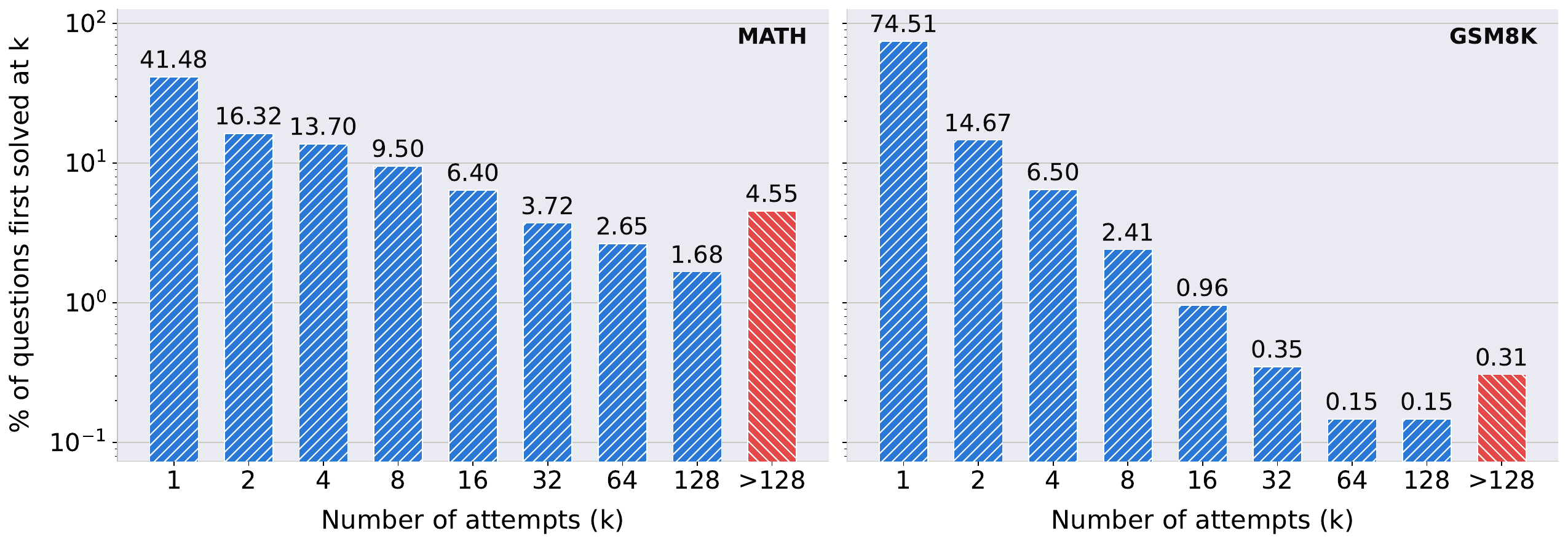}
    \caption{Question difficulty distributions for the MATH and GSM8K training datasets, measured by the number of rollout generations required to obtain the first correct solution.}
    \label{MotiveFigure}
\end{figure}

Despite its effectiveness, GRPO allocates a uniform rollout budget to all questions in a batch without accounting for question difficulty. This limitation is particularly important because a question contributes to the policy update only through the advantages computed from its rollout rewards. For relatively easy questions, all generated rollouts may yield the correct answer, whereas for excessively difficult questions, all generated rollouts may produce incorrect answers. In both cases, identical rewards yield zero advantages, resulting in no learning signal. Allocating additional rollouts to these questions can increase the likelihood of generating both correct and incorrect responses, leading to informative advantages. In contrast, moderately difficult questions typically yield sufficiently informative advantages with fewer rollouts. Thus, reallocating rollout budget from questions that already provide sufficient learning signals to those that require additional exploration has the potential to improve training efficiency by enabling a larger fraction of the training questions to effectively contribute to the policy update.


To illustrate the heterogeneity in question difficulty, Figure~\ref{MotiveFigure} shows the percentage of questions in the MATH~\citep{hendrycks2021measuring} and GSM8K~\citep{GSM8K} training sets that are first solved after the $k$-th rollout generated by Llama3.2-3B-Instruct. As observed, the number of rollouts required to obtain the first correct solution varies substantially across questions.
For example, while many questions of MATH are solved with a single rollout ($k=1$), more than $28\%$ require at least eight rollouts before producing a
reasoning trajectory with the correct final answer. This 
variation suggests that different questions 
require different rollout budgets, making a uniform allocation potentially inefficient. It 
thus raises the following key question: \textit{How should a fixed rollout budget be allocated across questions to maximize the effectiveness of GRPO training?}


\begin{figure}[t]
\centering
\includegraphics[width=\textwidth]{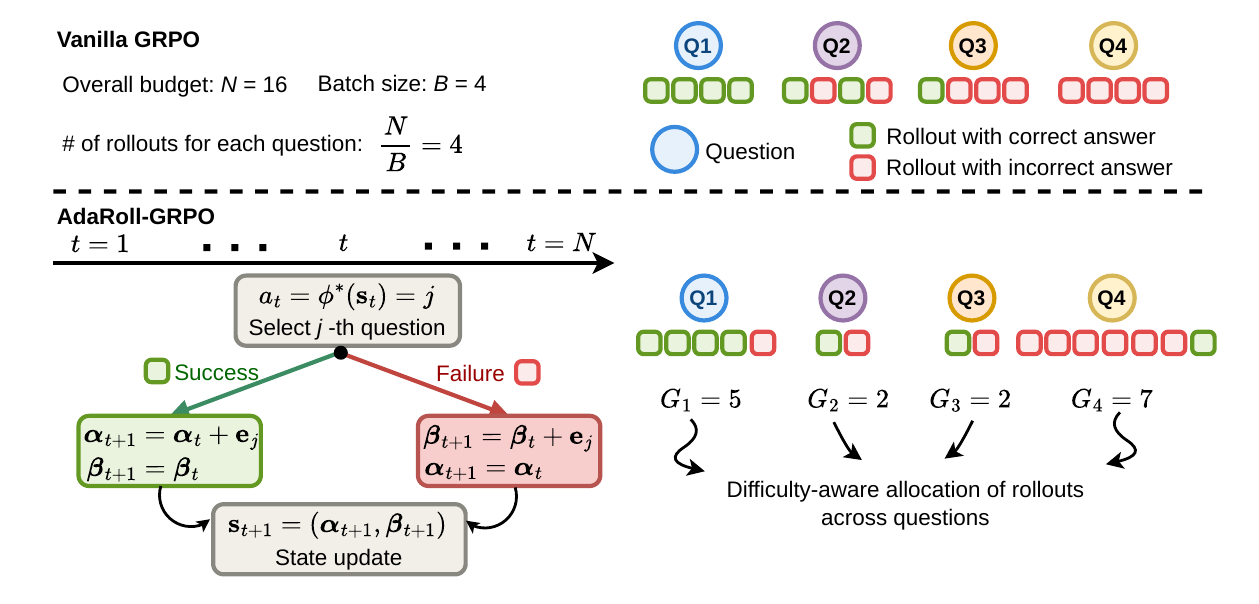} 
\caption{\textbf{Illustration of AdaRoll-GRPO.} In vanilla GRPO, for a batch of four questions with a total rollout budget of 16, four rollouts are uniformly allocated to each question. For some questions (e.g., Q1 and Q4), all rollouts may receive identical rewards (all 1s or all 0s), resulting in zero advantage and thus, no contribution to training. AdaRoll-GRPO instead learns a rollout allocation policy, $\phi^\ast$, which sequentially selects the next question to sample based on the current state. In this example, the learned policy reduces the rollout budget of Q2 and Q3, which already produce diverse rewards with only two rollouts each, and instead, allocates extra budget to Q1 and Q4 (5 and 7 rollouts, respectively). The additional rollouts increase the likelihood of obtaining diverse rewards for Q1 and Q4, producing non-zero advantages and making these questions effective training samples.}
\label{fig:overview}
\end{figure}

In this work, we propose \textbf{AdaRoll-GRPO}, an \underline{Ada}ptive \underline{Roll}out budget allocation framework for GRPO training, as illustrated in Figure~\ref{fig:overview}. AdaRoll-GRPO dynamically allocates the rollout budget across questions in each GRPO training iteration to improve training performance while maintaining the same total number of generated rollouts as the predefined rollout budget.

Specifically, we formulate the rollout budget allocation problem as an integer optimization problem~\citep{schrijver1998theory}, where the objective function captures different training utility functions, and the constraint ensures that the total number of generated rollouts remains within the given budget. However, this formulation leads to a challenging \textit{online} optimization problem~\citep{borodin2005online}, as the utility objective depends on the rewards of generated rollouts, which are unknown before rollout generation and can only be observed after allocation decisions are made.

Our contributions are summarized as follows:


\begin{itemize}


\item To address the online nature of the rollout allocation problem, we decompose it into a sequence of decision-making steps by modeling it as a finite-horizon Markov Decision Process (MDP)~\citep{puterman2014markov}, where the action at each step is to select a question for rollout generation.

\item The transition probabilities of the formulated MDP depend on the probability of the LLM generating a reasoning trajectory with a correct final answer, which is unknown during training. To account for this uncertainty, we formulate the problem using the Bayes-Adaptive MDP (BAMDP) ~\citep{BAMDPduff2002optimal, BAMDPduff2003design} framework, where the transition probabilities are represented through belief distributions and incorporated into an augmented belief state. We then solve the BAMDP using value iteration and derive a closed-form Bellman update for the value function.

\item The proposed value iteration procedure can be performed offline, and the resulting optimal allocation policy can be stored and directly applied during GRPO training, introducing no additional online optimization overhead.

\item Extensive experiments demonstrate that AdaRoll-GRPO effectively improves the training performance of different GRPO variants across various datasets and LLMs.

\end{itemize}



\section{Related Work}

Recent works have explored improving GRPO from two main perspectives: (i) modifying the GRPO optimization objective, and (ii) improving advantage estimation.

\subsection{Objective Modifications}

Several works enhance GRPO by modifying its optimization objective. DAPO~\citep{yu2026dapo} introduces asymmetric clipping and removes the KL divergence constraint to improve exploration and policy update effectiveness. Dr. GRPO~\citep{li2025drgrpo} removes the per-response length normalization from the 
objective, which biases model updates by upweighting shorter correct responses and downweighting longer incorrect ones. GSPO~\citep{zheng2025gspo} and TIC-GRPO~\citep{pang2025ticgrpo} replace token-level importance ratios with sequence-level ratios, treating the generated response as the optimization unit rather than individual tokens.
We focus on rollout budget allocation, which is orthogonal to these objective 
changes and can be combined with 
them to further improve 
training.

\subsection{Advantage Estimation Improvements}

A more specialized line of work focuses on improving the advantage estimation procedure in GRPO. In addition to modifying the objective, Dr. GRPO~\citep{li2025drgrpo} also identifies the difficulty bias introduced by reward standard deviation normalization and removes this normalization to mitigate the resulting bias in advantage estimation. 
MC-GRPO~\citep{mcgrpo2025} replaces mean-based reward centering with median-based centering to improve robustness under small rollout groups. KRPO~\citep{wang2025krpo} incorporates a Kalman filter to estimate the reward mean and variance used for advantage normalization, reducing the impact of noisy reward statistics in small rollout groups. MS-GRPO~\citep{wu2025msgrpo} constructs subgroups within each rollout group, computes relative advantages within each subgroup, and aggregates them to produce more reliable advantage estimates. 
In contrast, our work focuses on allocating rollout budget across questions rather than modifying advantage estimation. Thus, our approach is complementary to these methods and can be combined with different advantage estimation techniques.

\section{Background on GRPO}

Group Relative Policy Optimization (GRPO) \citep{shao2024deepseekmath, guo2025deepseek} was introduced as a 
an efficient alternative to Proximal Policy Optimization (PPO) \citep{schulman2017proximal} for LLM post-training. PPO relies on a separate value model to estimate the advantage function. 
Yet, training this additional model incurs extra computational overhead, and maintaining it requires additional memory resources. GRPO eliminates the need for the value model by estimating the advantage of each generated response based on the relative rewards of multiple sampled responses for the same input prompt. Specifically, instead of evaluating each response using a learned value function, GRPO compares responses generated for the same prompt and leverages their normalized rewards to determine which responses should be reinforced or discouraged during policy optimization.
For each input prompt $x_j$, GRPO generates a group of $G$ rollout responses:
\begin{align}\label{RolloutGeneration}
\mathcal{Y}(x_j)
=
\left\{
y_j^1,y_j^2,\ldots,y_j^G
\right\},
\end{align}
where $y_j^i$ denotes the $i$-th rollout generated for prompt $x_j$. Each rollout consists of a sequence of tokens as
$
y_j^i
=
\big(
y_j^{i,1},
y_j^{i,2},
\ldots,
y_j^{i,|y_j^i|}
\big),
$
where $y_j^{i,t}$ represents the $t$-th token of rollout $y_j^i$, and $y_j^{i,<t}$ denotes the preceding tokens before the $t$-th token.

Following PPO, GRPO uses a clipped surrogate objective:
\begin{align}\label{GRPO-Objective}
J(\theta)
=
\mathbb{E}_{x_j \sim \mathcal{D}}
\Bigg[
\frac{1}{G}
\sum_{i=1}^{G}
\frac{1}{|y_j^i|}
\sum_{t=1}^{|y_j^i|}
\min
\Bigg(
\rho_j^{i,t}(\theta)
A_j^{i,t},\,
\mathrm{clip}
\Big[
\rho_j^{i,t}(\theta),
1-\epsilon,
1+\epsilon
\Big]
A_j^{i,t}
\Bigg)
\Bigg],
\end{align}
where $\rho_j^{i,t}(\theta)$ denotes the importance ratio for the $t$-th token of the $i$-th rollout, defined as
\begin{align}
\rho_j^{i,t}(\theta)
=
\frac{
\pi_{\theta}
\left(
y_j^{i,t}
\mid
x_j,
y_j^{i,<t}
\right)
}{
\pi_{\mathrm{old}}
\left(
y_j^{i,t}
\mid
x_j,
y_j^{i,<t}
\right)
}.
\end{align}
Here, $\pi_{\theta}(y_j^{i,t} \mid x_j,y_j^{i,<t})$ denotes the probability assigned by 
$\pi_{\theta}$, the current policy being optimized, to token $y_j^{i,t}$ given the prompt $x_j$ and preceding tokens $y_j^{i,<t}$, and $\pi_{\mathrm{old}}$ denotes the reference policy used to generate the rollouts.

Unlike PPO, where the advantage is estimated using a learned value model, GRPO computes the advantage using the relative rewards within each rollout group. Specifically, the advantage of $t$-th token in rollout $y_j^i$ is calculated as
\begin{align}
A_j^{i,t}
=
\frac{
r_j^i-\mathrm{mean}\{r_j^1,r_j^2,\ldots,r_j^G\}
}{
\mathrm{std}\{r_j^1,r_j^2,\ldots,r_j^G\}
},
\qquad \forall t,
\end{align}
where $r_j^i$ denotes the reward of rollout $y_j^i$. Hence, responses receiving higher rewards compared to other responses generated for the same prompt obtain positive advantages, while lower-quality responses receive negative advantages.

Throughout this work, we consider binary rewards, i.e.,
$r_j^i \in \{1, 0\},$ indicating whether the final answer produced by rollout $y_j^i$ is correct or not.

\section{Rollout Budget Allocation Problem Formulation}

As described in \eqref{RolloutGeneration}, GRPO generates an equal number of rollouts, denoted by $G$, for each question in the batch, which may be suboptimal. Questions at different difficulty levels can require different rollout budgets to yield informative reward signals: very easy and very hard questions may require additional rollouts to obtain diverse outcomes, while moderately difficult questions may already provide sufficient learning signals with fewer rollouts. Below, we provide an example to illustrate the limitation of uniform rollout budget allocation.


\subsection{Illustrative Example}
\begin{figure}[t]
\centering
\includegraphics[width=\textwidth]{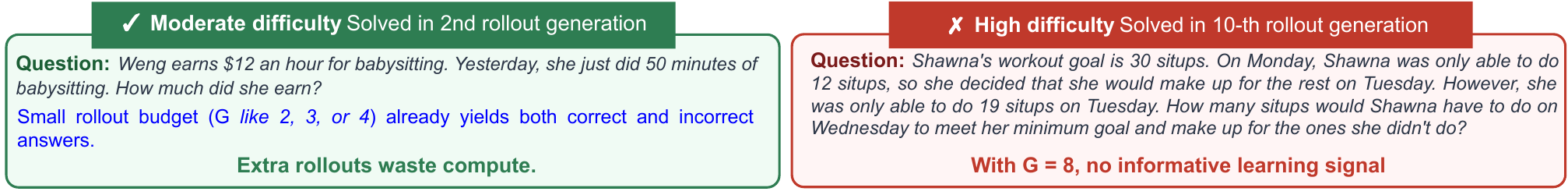} 
\caption{Illustration of the limitation of uniform rollout budget allocation in GRPO. With a fixed budget of $G=8$ rollouts per question, the harder question fails to generate any correct rollout, resulting in zero rewards for the entire group and no contribution to the policy update, while the easier question is allocated more rollouts than necessary.}
\label{fig2-illustation}
\end{figure}

We provide an illustrative example to show the drawback of uniform rollout allocation.
We consider a batch of two mathematical reasoning questions from GSM8K~\citep{GSM8K}, shown in Figure~\ref{fig2-illustation}: the left question is moderately difficult and is solved by the base model (Llama3.2-3B-Instruct) after two rollout attempts, while the right question is more challenging and is first solved after ten rollout generations.

Assuming vanilla GRPO with $G=8$, the harder question fails to generate any correct reasoning trajectory within the allocated rollout budget. Hence, all rewards in its rollout group are zero, yielding zero advantages and preventing the sample from contributing to training. In contrast, the moderately difficult question produces both successful and unsuccessful trajectories within only a few rollouts, providing informative reward signals. Thus, a more effective strategy could allocate a non-uniform rollout budget across questions, for example, $G_1=4$ for the easier question and $G_2=12$ for the harder question, allowing the harder question to obtain informative rewards while avoiding unnecessary rollouts for the question that already provides a sufficient learning signal.


This example highlights the need for a judicious rollout allocation strategy instead of uniform allocation, motivating the question: \textit{How should rollout budgets be allocated to most effectively improve the learning process?}




\subsection{Problem Formulation}
Throughout this work, we refer to the standard GRPO with a uniform rollout budget allocation of $G$ rollouts for every question as the vanilla GRPO method. In each training iteration of GRPO, given a training batch $\{x_1, x_2, \ldots, x_B\}$, we aim to design a rollout allocation vector $\mathbf{G} \triangleq (G_1, G_2, \ldots, G_B) \in \mathbb{N}^{B}$ to improve the training process. 

For question $j$, let
\(
\mathbf{r}_j(G_j) = \big(r_j^1, \ldots, r_j^{G_j}\big) \in \{0,1\}^{G_j}
\)
denote the ordered tuple of rewards obtained from its $G_j$ allocated rollouts, where $\mathbf{r}_j(G_j)$ is a function of the allocated rollout budget $G_j$. Let
\(
\mathbf{r}(\mathbf{G}) \triangleq \big(\mathbf{r}_1(G_1), \ldots, \mathbf{r}_B(G_B)\big)
\)
denote the collection of rollout rewards for all questions in the batch. Since rollout rewards directly affect the policy update and, hence, the training performance, we define the training utility as
$ U\big(\mathbf{r}(\mathbf{G})\big),
$
which is a function of the rollout rewards $\mathbf{r}(\mathbf{G})$. We present several instances of this utility function later. Since the rollout rewards are stochastic, the objective is to maximize the expected training utility over the reward outcomes. The rollout budget allocation problem is 
formulated as
\begin{subequations}\label{original_problem}
    \begin{align}
        \max_{\mathbf{G} \in \mathbb{N}^B} \quad & \mathbb{E}_{\mathbf{r}(\mathbf{G})} \Big[U\big(\mathbf{r}(\mathbf{G})\big) \Big] \label{original_problem_objective} \\
        \text{s. t.} \quad & \sum_{j=1}^{B} G_j = N, \label{original_problem:C1}
    \end{align}
\end{subequations}
where constraint \eqref{original_problem:C1} ensures that the total number of allocated rollouts across all questions remains equal to the overall rollout budget $N=B \cdot G$ used by vanilla GRPO. 

The choice of the utility function $U\big(\mathbf{r}(\mathbf{G})\big)$ determines the objective in \eqref{original_problem} and consequently the resulting rollout allocation strategy. Below, we consider several choices of utility functions designed to improve training quality.

\paragraph{Minimum reward variance ($U_1$)}
A larger diversity of rewards among rollouts generated for a question results in more non-zero advantage values, which provide more effective training signals. The reward diversity of a question is measured by the variance of its rollout rewards, given by
\begin{align}
    \mathrm{Var}_j
    =
    \frac{1}{G_j}
    \sum_{i=1}^{G_j}
    \left(
    r_j^i-\mathrm{mean}\{r_j^1,r_j^2,\ldots,r_j^{G_j}\}
    \right)^2.
\end{align}

To ensure that all questions in the batch provide diverse reward signals, we define the training utility function as the minimum reward variance across all questions:
\begin{align}\label{Min-var-utility}
    U_1\big(\mathbf{r}(\mathbf{G})\big)
    \triangleq
    \min_{1\le j \le B} \mathrm{Var}_j.
\end{align}

\paragraph{Minimum advantage contribution ($U_2$)}
Since the GRPO objective in \eqref{GRPO-Objective} scales with the magnitude of advantage $|A_j^{i,t}|$, the advantage magnitude directly influences the contribution of each rollout to the policy update. Rollouts with larger advantages contribute more to the batch gradient and can dominate the optimization process. 
Thus, we measure the contribution of each question to training using the sum of squared advantages of its generated rollouts: $\sum_{i=1}^{G_j} \left|A_j^i\right|^2 .$

To ensure that all questions in the batch contribute effectively to training, we can define the utility function as the minimum advantage contribution across all questions:
\begin{align}\label{Min-sum-sq_advantage-utility}
    U_2\big(\mathbf{r}(\mathbf{G})\big)
    \triangleq
    \min_{1\le j\le B}
    \sum_{i=1}^{G_j}
    \left|A_j^i\right|^2 .
\end{align}

\paragraph{Summation of maximum rewards ($U_3$)}
To provide an effective learning signal for each question, the model should generate at least one correct reasoning trajectory from its rollouts. Therefore, we consider the maximum reward among the generated rollouts for each question as an indicator of whether that question contributes to the training process. Specifically, if
$\max_{1\le i \le G_j} r_j^i = 1$,
then at least one correct reasoning trajectory is generated for question $j$, allowing the training process to benefit from this sample.

To account for the entire batch, we define the utility function as the sum of the maximum rewards over questions:
\begin{align}\label{Sum-max-R-utility}
    U_3\big(\mathbf{r}(\mathbf{G})\big)
    \triangleq
    \sum_{j=1}^{B}
    \max_{1\le i\le G_j} r_j^i .
\end{align}

Appendix~\ref{Appendix:alternative-objectives} details three additional objectives: minimum of maximum rewards, summation of reward variances, and summation of advantage contributions.

\section{Adaptive Rollout Budget Allocation}

The rollout budget allocation problem in \eqref{original_problem} with the utility functions defined in \eqref{Min-var-utility}, \eqref{Min-sum-sq_advantage-utility}, and \eqref{Sum-max-R-utility} is an \textit{online} integer programming problem. In particular, the objective value of each utility function depends on the rewards of the generated rollouts, which are only observable after allocating the rollout budget and generating the corresponding responses. 
To address this online decision-making problem, we formulate it as a Markov Decision Process (MDP), described below.

\subsection{MDP Formalization}\label{MDP-Formalization}

In problem \eqref{original_problem}, the total number of generated rollouts is limited to $N$. Therefore, we model the rollout budget allocation process as a finite-horizon MDP with horizon length $N$, where each decision step corresponds to allocating one rollout to one of the questions in the batch. Specifically, we denote the decision step by $t$, where $1\le t\le N$. 

\paragraph{MDP action}
At step $t$, the action $a_t \in \{1,\ldots,B\}$ selects the index of a question in the batch for which a new rollout is generated.

\paragraph{MDP state} The state $\mathbf{s}_t$ represents the observed reward information at step $t$, after $t-1$ rollout allocations, and is defined as
\begin{align}\label{MDP_state}
    \mathbf{s}_t = \Big( (n_{1,t}, n_{1,t}^s),(n_{2,t}, n_{2,t}^s),\ldots, (n_{B,t}, n_{B,t}^s) \Big),
\end{align}
where $n_{j,t}$ denotes the total number of rollouts allocated to question $j$ up to step $t$, and $n_{j,t}^s$ denotes the number of those rollouts that produce correct answers.
Note that the training utility functions in \eqref{Min-var-utility}--\eqref{Sum-max-R-utility} (and \eqref{Min-max-R-utility}--\eqref{Sum-sum-sq_advantage-utility} in Appendix~\ref{Appendix:alternative-objectives}) depend only on the total number of rollouts generated for each question and the numbers of successful and unsuccessful rollouts, all of which are fully captured by our state definition.

\paragraph{MDP state transition}

Given the current state $\mathbf{s}_t$ and selected action $a_t=j$, a new rollout is generated for question $j$, and its reward is observed. The state then transitions to $\mathbf{s}_{t+1}$ by updating only the tuple $(n_{j,t}, n_{j,t}^s)$ corresponding to question $j$, with all other components remaining unchanged:
\begin{align}\label{MDP-transition-prob}
\big(n_{j,t+1}, n_{j,t+1}^s \big)
=
\begin{cases}
\big(n_{j,t}+1,\, n_{j,t}^s+1\big), & \text{with probability } p_j, \\
\big(n_{j,t}+1,\, n_{j,t}^s\big), & \text{with probability } 1-p_j,
\end{cases}
\end{align}
where $p_j$ denotes the probability that the LLM generates a rollout with a correct final answer for question $x_j$.

\paragraph{MDP immediate reward}

Since the goal of the rollout budget allocation problem is to maximize the final utility function $U\big(\mathbf{r}(\mathbf{G})\big)$, we define the immediate reward to be zero for all intermediate decision steps and assign the utility value only after all $N$ rollout allocations have been performed:
\begin{align}\label{mdp-immediater}
    r_t^{\mathrm{MDP}} =
    \begin{cases}
        0, & t=1,\ldots,N-1,\\
        U\big(\mathbf{r}(\mathbf{G})\big), & t=N,
    \end{cases}
\end{align}
where $\mathbf{G}=(G_1,\ldots,G_B)$ is the final rollout allocation vector obtained after the $N$ sequential allocation decisions.

\begin{remark}
We note that the MDP defined above generalizes Problem~\eqref{original_problem} by allowing the rollout allocation to be determined sequentially based on the observed reward history, rather than committing to a fixed allocation vector before observing any rewards. The finite horizon of length $N$ enforces the rollout budget constraint in~\eqref{original_problem:C1}, while the expected cumulative reward of the MDP under any policy equals the expected utility achieved under the same policy.
\end{remark}

\subsection{Bayes-Adaptive MDP (BAMDP) Formalization}

During GRPO training, the model is updated in each iteration, causing the probability $p_j$ that it successfully solves question $j$ to change per iteration. Hence, the transition probabilities of the MDP in \eqref{MDP-transition-prob} are unknown during each training iteration. To account for this uncertainty in the transition dynamics, we leverage the Bayes-Adaptive MDP (BAMDP) framework, which extends MDPs to settings with unknown transition dynamics by maintaining a belief distribution over the unknown transition parameters~\citep{BAMDPduff2002optimal, BAMDPduff2003design}.


We model the unknown success probability $p_j$ as a random variable with a Beta prior. Due to the conjugacy between the Beta distribution and the Bernoulli likelihood, the posterior belief over $p_j$ also follows a Beta distribution after observing rollout outcomes~\citep{murphy2012machine}.

\paragraph{Prior belief over $p_j$}

We initialize the belief over $p_j$ with a Beta prior, \(p_j \sim \mathrm{Beta}(\alpha_j^1,\beta_j^1),\)
where $\alpha_j^1=\beta_j^1=1$ $\forall j$. This prior is equivalent to a uniform distribution over $[0,1]$, reflecting the absence of prior knowledge about the probability that the model successfully solves question $j$ before observing any rollout outcomes.



\paragraph{BAMDP state}

In the BAMDP, a belief distribution is maintained over the unknown transition parameters. The state is therefore augmented with the parameters of this distribution, forming the augmented belief state.
Specifically, at step $t$, we define the BAMDP state as $
    \mathbf{s}_t = (\boldsymbol{\alpha}_t,\boldsymbol{\beta}_t), $
where $
    \boldsymbol{\alpha}_t
    \triangleq
    (\alpha_1^t,\ldots,\alpha_B^t), 
    \boldsymbol{\beta}_t
    \triangleq
    (\beta_1^t,\ldots,\beta_B^t).$
Here, $\alpha_j^t$ and $\beta_j^t$ denote the success and failure parameters of the belief distribution over $p_j$ for question $j$, respectively.

Note that the state consists only of the belief parameters $\boldsymbol{\alpha}_t$ and $\boldsymbol{\beta}_t$ and does not include the original MDP state in \eqref{MDP_state}. As shown later, these belief parameters fully capture the information contained in the MDP state, making it unnecessary to include the original state in the BAMDP state.

\paragraph{BAMDP state transition}

In step $t$, given the current state $\mathbf{s}_t=(\boldsymbol{\alpha}_t,\boldsymbol{\beta}_t)$, the belief over the success probability of question $j$ is $p_j \sim \mathrm{Beta}(\alpha_j^t,\beta_j^t)$.

The mean of the belief distribution provides an estimate of the success probability of question $j$ at step $t$:
\begin{align}\label{transitionprobestim}
    \hat{p}_j^t
    =
    \frac{\alpha_j^t}
    {\alpha_j^t+\beta_j^t}.
\end{align}

Suppose action $a_t=j$ is selected, an additional rollout is generated for question $x_j$, and its corresponding reward $r_t\in\{0,1\}$ is observed. The posterior belief over $p_j$ becomes
\begin{align}\label{posterior-update}
    p_j
    \sim
    \mathrm{Beta}
    \left(
    \alpha_j^{t+1},
    \beta_j^{t+1}
    \right),
\end{align}
where $\alpha_j^{t+1}=\alpha_j^t+r_t$ and $\beta_j^{t+1}=\beta_j^t+1-r_t$.\footnote{The derivation of the posterior update is given in Appendix~\ref{AppendixA-PosteriorUpdate}.}
Therefore, the transition of the belief state can be written as
\begin{align}
(\boldsymbol{\alpha}_{t+1},\boldsymbol{\beta}_{t+1}) =
\begin{cases}
(\boldsymbol{\alpha}_t+\mathbf{e}_j,\;\boldsymbol{\beta}_t),
& \text{with probability } \hat{p}_j^t,\\
(\boldsymbol{\alpha}_t,\;\boldsymbol{\beta}_t+\mathbf{e}_j),
& \text{with probability } 1-\hat{p}_j^t,
\end{cases}
\end{align}
where $\mathbf{e}_j$ denotes the $j$-th standard basis vector.

We note that the posterior update in \eqref{posterior-update} implies that $\boldsymbol{\alpha}_t$ and $\boldsymbol{\beta}_t$ record the numbers of successful and unsuccessful rollouts for each question. Hence, they fully determine the original MDP state in \eqref{MDP_state}, allowing the BAMDP state to be represented solely by the belief parameters.


\section{Optimal Rollout Allocation Policy}

\begin{algorithm}[tb]
\caption{Rollout Budget Allocation via Value Iteration}
\label{alg:value_iteration}
\small
\textbf{Input}: Batch size $B$, rollout budget $N$\\
\textbf{Output}: Optimal rollout allocation policy $\phi^{\ast}$
\begin{algorithmic}[1]
\STATE Initialize terminal value: $
V(\mathbf{s}_{N+1}) = r_N^{\mathrm{MDP}}.$
\FOR{$t=N$ \TO $1$}

    \FOR{each belief state $\mathbf{s}_t=(\boldsymbol{\alpha}_t,\boldsymbol{\beta}_t)$}

        \STATE Estimation of success probability: $\hat{p}_j^t =
    \frac{\alpha_j^t}
    {\alpha_j^t+\beta_j^t}, \forall j.$
        \STATE Compute the value of selecting each question:
        \begin{align}
        Q(\mathbf{s}_t,j)
        &= \hat{p}_j^t
        V(\boldsymbol{\alpha}_t+\mathbf{e}_j,\boldsymbol{\beta}_t)
        +
        (1-\hat{p}_j^t)
        V(\boldsymbol{\alpha}_t,\boldsymbol{\beta}_t+\mathbf{e}_j).\nonumber
        \end{align}

        \STATE Update value: $
        V(\mathbf{s}_t)
        =
        \max_j Q(\mathbf{s}_t,j)$.

        \STATE Store optimal action:
        $
        \phi^{\ast}(\mathbf{s}_t)
        =
        \arg\max_j Q(\mathbf{s}_t,j)$.

    \ENDFOR

\ENDFOR
\STATE \textbf{return} optimal policy $\phi^{\ast}$
\end{algorithmic}
\end{algorithm}

To solve the formulated BAMDP, we use the value iteration
algorithm~\citep{sutton2018reinforcement}. Specifically, we define the value function
$V(\mathbf{s}_t)$ as the expected utility obtained by following the
optimal allocation policy $\phi^{\ast}$ from state $\mathbf{s}_t$
until the end of the decision horizon. 
Recall from~\eqref{mdp-immediater} that the reward
$r_N^{\mathrm{MDP}}$ is realized only after the $N$-th rollout is
generated, i.e., upon reaching state $\mathbf{s}_{N+1}$. Since no
further decisions are made after this point, the value of the
terminal state is simply this reward: $V(\mathbf{s}_{N+1}) = r_N^{\mathrm{MDP}}. $
For intermediate steps, using the Bellman optimality equation~\citep{bellman1957dynamic, sutton2018reinforcement}, we have
\begin{align}
V(\boldsymbol{\alpha}_t,\boldsymbol{\beta}_t)
=
\max_{j\in\{1,\ldots,B\}}
\Big[
\hat{p}_j^t\,
V(\boldsymbol{\alpha}_t+\mathbf{e}_j,\boldsymbol{\beta}_t)
+
(1-\hat{p}_j^t)\,
V(\boldsymbol{\alpha}_t,\boldsymbol{\beta}_t+\mathbf{e}_j)
\Big].
\end{align}
The first term corresponds to the case where the generated rollout for question $j$ is successful, and the belief parameter $\alpha_j^t$ is incremented. The second term represents the unsuccessful case, where the failure parameter $\beta_j^t$ is incremented. Algorithm~\ref{alg:value_iteration} summarizes the value iteration procedure.

\paragraph{Optimal policy extraction}

Since value iteration is independent of the GRPO training iterations, it is performed once offline before training. The resulting optimal policy is stored as a lookup table mapping each belief state to the next question to sample. During training, rollout allocation is therefore reduced to a simple table lookup, incurring no additional online computational cost.

\paragraph{Computational complexity}

The offline value iteration has complexity $O(BN^{2B})$, where $N^{2B}$ is the number of reachable belief states and each state requires evaluating $B$ 
actions. For a fixed batch size, this complexity is polynomial in the rollout budget $N$. As it is incurred only once offline, it does not affect the per-step computational cost of GRPO training.










\section{Experiments}

We evaluate AdaRoll-GRPO on four benchmark datasets spanning three reasoning domains: mathematical reasoning, logical reasoning, and commonsense reasoning. Experiments are conducted using Llama 3.1-8B-Instruct~\citep{llama32024}, Llama 3.2-3B-Instruct~\citep{llama32024}, Gemma 3-4B-IT~\citep{gemma3_technical_report}, and Qwen 2-MATH-1.5B~\citep{yang2024qwen2}. 
The implementation of AdaRoll-GRPO and all baseline methods is available in the GitHub repository linked in the following footnote.\footnote{\url{https://github.com/faezemoradik/AdaRollGRPO-Release}}

\subsection{Datasets and Tasks}

We evaluate mathematical reasoning on GSM8K~\citep{cobbe2021training} and MATH~\citep{hendrycks2021measuring}, logical reasoning on LogiQA~\citep{liu2021logiqa}, and commonsense reasoning on CoLoTa~\citep{toroghi2025colota}. Additional details on the datasets are provided in Appendix~\ref{Appendix:taskData}.

\subsection{Baselines and Evaluation}

The proposed rollout allocation strategy does not modify the underlying GRPO objective and can therefore be integrated with any GRPO variant. We evaluate it using vanilla GRPO, Dr.GRPO~\citep{li2025drgrpo}, and DAPO~\citep{yu2026dapo}. We compare each variant with its counterpart augmented by the proposed rollout allocation strategy to isolate the effect of rollout allocation.

Each trained model is evaluated on the corresponding test set of its training dataset using exact-match accuracy, where a prediction is considered correct only if the generated final answer exactly matches the ground-truth answer.

\begin{myrq}\label{RQ1}
\textbf{How do different training utility objectives affect the final performance of AdaRoll-GRPO?}
\end{myrq}
To evaluate the impact of different training utility functions, we equip vanilla GRPO with rollout allocation policies learned using each of the six utility functions defined in \eqref{Min-var-utility}--\eqref{Sum-max-R-utility} and \eqref{Min-max-R-utility}--\eqref{Sum-sum-sq_advantage-utility} (given in Appendix~\ref{Appendix:alternative-objectives}). We then compare the final accuracy of the resulting models on the test split of each dataset. 
Table~\ref{tab:adaroll_variants_results} reports the results, with the columns ordered according to the utility functions listed above.

As shown, although no single objective consistently outperforms the others across all evaluated settings, the minimum reward variance objective, $U_1$, achieves the strongest overall performance, attaining the best results in 6 of the 14 benchmark-model combinations. We therefore adopt the minimum reward variance objective in the remaining experiments. 
More detailed results and analyses of how different training utility functions affect the reward dynamics during training, and how they shape the learned rollout allocation policy, which in turn determines the distribution of rollouts across questions within a batch, are provided in Appendix~\ref{Additional_Experimnetal_analysis}.


\begin{table}[t]
\caption{Final test accuracy ($\%$) of models trained with AdaRoll-GRPO using different training utility functions.}
\label{tab:adaroll_variants_results}
\centering
\small
\setlength{\tabcolsep}{4pt}
\renewcommand{\arraystretch}{1.15}
\begin{tabular}{lcccccc}
\toprule
\textbf{Dataset} &
$U_1$ &
$U_2$ &
$U_3$ &
$U_4$ &
$U_5$ &
$U_6$ \\
\midrule
\midrule
\addlinespace[2pt]
\multicolumn{7}{c}{\textbf{Llama 3.1-8B-Instruct}}\\
\addlinespace[2pt]
MATH   & \textbf{48.60} & 44.00 & 47.00 & 45.00 & 47.60 & 46.80 \\
GSM8K  & 86.05 & \textbf{87.19} & 86.13 & 85.06 & 86.13 & 85.75 \\
CoLoTa & 58.24 & 54.71 & 54.71 & 57.06 & 60.00 & \textbf{60.59} \\
LogiQA & 43.93 & 43.01 & \textbf{45.16} & 41.47 & 43.78 & 41.78 \\
\addlinespace[5pt]
\midrule
\midrule
\addlinespace[2pt]
\multicolumn{7}{c}{\textbf{Llama 3.2-3B-Instruct}}\\
\addlinespace[2pt]
MATH   & \textbf{39.60} & 37.80 & 37.80 & 36.40 & 40.60 & 40.80 \\
GSM8K  & 75.21 & 75.21 & \textbf{75.36} & 73.77 & 74.83 & 74.60 \\
CoLoTa & \textbf{58.82} & 57.65 & 56.47 & 54.71 & 58.24 & 57.65 \\
LogiQA & \textbf{37.02} & 36.41 & 36.41 & 36.25 & 36.71 & 34.72 \\
\addlinespace[5pt]
\midrule
\midrule
\addlinespace[2pt]
\multicolumn{7}{c}{\textbf{Gemma 3-4B-IT}}\\
\addlinespace[2pt]
MATH   & 68.00 & 64.60 & 67.80 & 66.20 & \textbf{69.00} & 65.40 \\
GSM8K  & \textbf{88.86} &  87.49 & 88.40 & 86.81 & 87.04 & 87.19 \\
CoLoTa & \textbf{77.65} & 65.88 & 70.00 & 66.47 & 71.18 & 70.59\\
LogiQA & 45.16 & 46.24 & \textbf{46.54} & 42.24 & 44.85 & 43.47 \\
\addlinespace[5pt]
\midrule
\midrule
\addlinespace[2pt]
\multicolumn{7}{c}{\textbf{Qwen 2-Math-1.5B}}\\
\addlinespace[2pt]
MATH   & 57.40 & 56.20 & 54.40 & 54.20 & \textbf{58.00} & 55.60 \\
GSM8K  & 77.94 & 76.42 & 77.63 & \textbf{81.12} & 79.08 & 79.23 \\
\bottomrule
\end{tabular}
\end{table}

\begin{myrq}\label{RQ2}
\textbf{How does the proposed rollout allocation strategy affect the final training performance of vanilla GRPO, Dr.GRPO, and DAPO?
}
\end{myrq}

Table~\ref{tab:adaroll_results} reports the final test accuracy achieved by each GRPO variant under two rollout allocation strategies: The \textbf{Def.} subcolumn corresponds to the original GRPO variant, whereas the \textbf{+ Ada.} subcolumn reports the performance of its AdaRoll-enhanced counterpart.
Notably, augmenting each baseline with the proposed rollout allocation strategy consistently improves performance over its original counterpart. Furthermore, comparing the vanilla GRPO results in this table with those in Table~\ref{tab:adaroll_variants_results} shows that AdaRoll-GRPO with the other training utility functions also outperforms vanilla GRPO in the majority of cases. These improvements show that adaptive rollout allocation provides a robust and generalizable approach for enhancing GRPO-based training.








\begin{table}[t]
\caption{Comparison of the final test accuracy ($\%$) of GRPO variants and their AdaRoll-enhanced counterparts.
}
\label{tab:adaroll_results}
\centering
\small
\setlength{\tabcolsep}{4pt}
\renewcommand{\arraystretch}{1.15}

\begin{tabular}{lcccccc}
\toprule
&
\multicolumn{2}{c}{\textbf{Vanilla GRPO}} &
\multicolumn{2}{c}{\textbf{Dr. GRPO}} &
\multicolumn{2}{c}{\textbf{DAPO}} \\
\cmidrule(lr){2-3}
\cmidrule(lr){4-5}
\cmidrule(lr){6-7}
\textbf{Dataset} &
\textbf{Def.} &
\textbf{+ Ada.} &
\textbf{Def.} &
\textbf{+ Ada.} &
\textbf{Def.} &
\textbf{+ Ada.} \\
\midrule
\midrule

\addlinespace[2pt]
\multicolumn{7}{c}{\textbf{Llama 3.1-8B-Instruct}}\\
\addlinespace[2pt]

MATH   & 42.60 & \textbf{48.60} & 44.40 & \textbf{48.80} & 46.60 & \textbf{49.40} \\
GSM8K  & 83.02 & \textbf{86.05} & 82.56 & \textbf{84.08} & 83.32 & \textbf{86.35} \\
CoLoTa & 50.00 & \textbf{58.24} & 46.47 & \textbf{53.52} & 48.23 & \textbf{54.12} \\
LogiQA & 41.17 & \textbf{43.93} & 42.70 & \textbf{44.85} & 43.47 & \textbf{46.24} \\

\addlinespace[5pt]
\midrule
\midrule
\addlinespace[2pt]

\multicolumn{7}{c}{\textbf{Llama 3.2-3B-Instruct}}\\
\addlinespace[2pt]

MATH   & 35.60 & \textbf{39.60} & 36.60 & \textbf{38.40} & 37.20 & \textbf{39.80} \\
GSM8K  & 73.54 & \textbf{75.21} & 72.71 & \textbf{74.37} & 73.01 & \textbf{74.98} \\
CoLoTa & 54.71 & \textbf{58.82} & 51.18 & \textbf{55.88} & 55.29 & \textbf{59.41} \\
LogiQA & 31.18 & \textbf{37.02} & 30.72 & \textbf{35.18} & 30.41 & \textbf{36.41} \\

\addlinespace[5pt]
\midrule
\midrule
\addlinespace[2pt]

\multicolumn{7}{c}{\textbf{Gemma 3-4B-IT}}\\
\addlinespace[2pt]

MATH   & 63.20 & \textbf{68.00} & 65.80 & \textbf{68.40} & 64.20 & \textbf{67.20} \\
GSM8K  & 86.42 & \textbf{88.86} & 87.04 & \textbf{89.31} & 86.28 & \textbf{88.86} \\
CoLoTa & 66.47 & \textbf{77.65} & 69.41 & \textbf{75.88} & 67.65 & \textbf{76.47} \\
LogiQA & 42.09 & \textbf{45.16} & 43.16 & \textbf{45.76} & 42.86 & \textbf{45.93} \\

\addlinespace[5pt]
\midrule
\midrule
\addlinespace[2pt]

\multicolumn{7}{c}{\textbf{Qwen 2-Math-1.5B}}\\
\addlinespace[2pt]

MATH   & 56.20 & \textbf{57.40} & 55.80 & \textbf{57.60} & 56.40 & \textbf{58.80} \\
GSM8K  & 77.79 & \textbf{77.94} & 77.94 & \textbf{79.08} &  76.50 & \textbf{78.09} \\
\bottomrule
\end{tabular}
\end{table}

\begin{figure}[!t]
    \centering
    \includegraphics[width=0.6\textwidth]{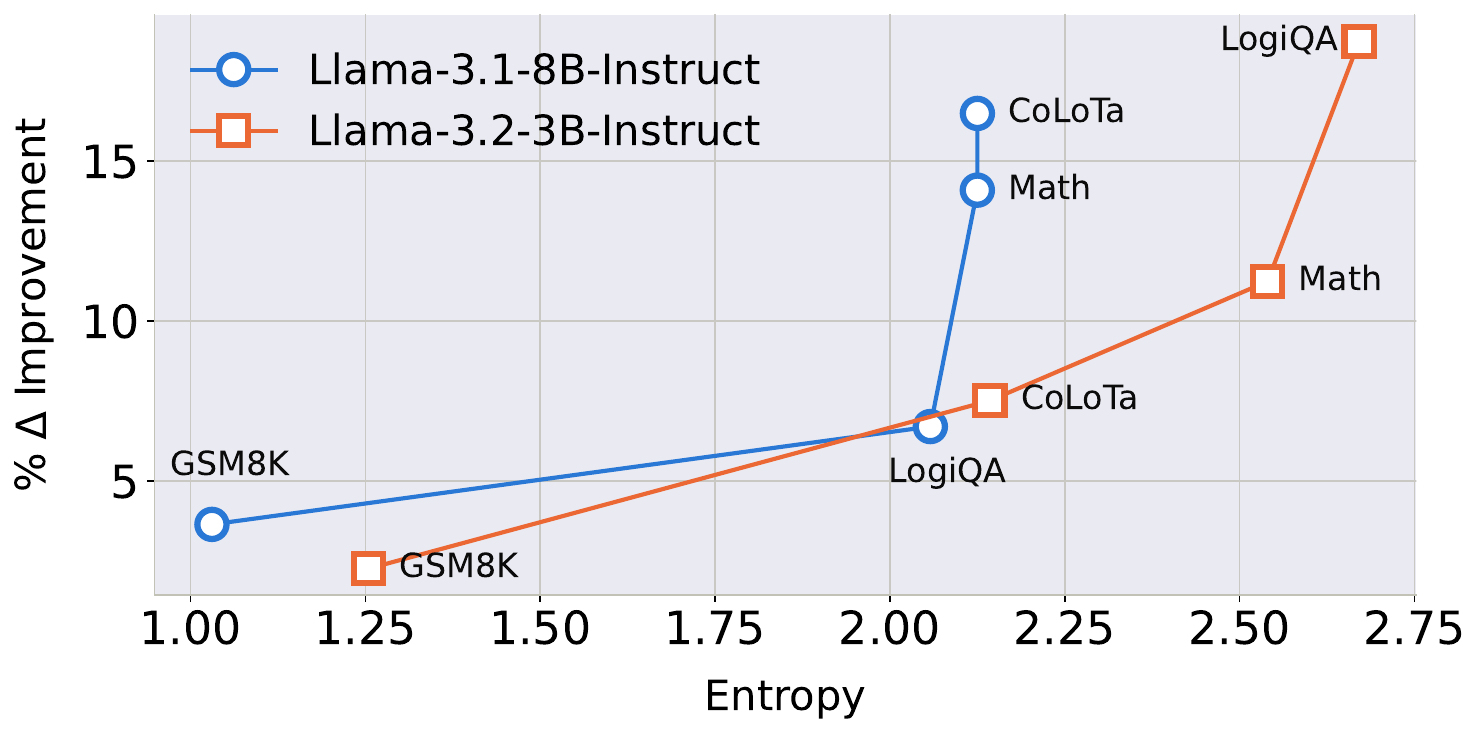}
    \caption{Normalized relative improvement of AdaRoll-GRPO vs. entropy of the dataset difficulty distributions.
    }
    \label{Delta_vs_entropy}
\end{figure}

\begin{myrq}\label{RQ3}
\textbf{
How does training data heterogeneity impact the effectiveness of AdaRoll-GRPO?}
\end{myrq}

To quantify the heterogeneity of each training dataset, we compute the entropy of its question difficulty distribution, where the difficulty of a question is measured by the number of rollout generations required by the model prior to training to obtain the first correct solution (similar to the distributions shown in Figure~\ref{MotiveFigure}).
A higher entropy indicates greater heterogeneity, as questions are more evenly distributed across different difficulty levels.

Figure~\ref{Delta_vs_entropy} shows the normalized relative improvement of AdaRoll-GRPO over vanilla GRPO against this entropy measure. As shown, the performance gain increases with dataset entropy. This suggests that adaptive rollout allocation is particularly beneficial for datasets with more diverse question difficulty, where allocating different rollout budgets to different questions is more effective than a uniform allocation.

\section{Conclusion}
We presented AdaRoll-GRPO, a framework for difficulty-aware rollout budget allocation in GRPO training. By formulating rollout allocation as a BAMDP, AdaRoll-GRPO learns an adaptive policy that allocates a fixed rollout budget across questions according to their estimated difficulty, without introducing additional online optimization overhead. Experiments across multiple reasoning benchmarks, LLM architectures, and variants of GRPO demonstrate that AdaRoll-GRPO consistently improves training performance under the same rollout budget. Overall, AdaRoll-GRPO provides an efficient and effective framework for improving GRPO-based LLM post-training through adaptive rollout allocation.


\label{maintextend}
\clearpage
\begingroup
\makeatletter
\let\@biblabel\NAT@biblabelnum
\let\@bibsetup\NAT@bibsetnum
\makeatother
\bibliography{references}
\bibliographystyle{iclr2027_conference}
\endgroup
\clearpage
\appendix
This appendix is organized as follows. Appendix~\ref{Appendix:alternative-objectives} introduces three additional utility functions beyond those presented in the main paper. Appendix~\ref{AppendixA-PosteriorUpdate} derives the posterior update equation presented in~\eqref{posterior-update}. Appendix~\ref{Appendix:taskData} provides additional details on the tasks and datasets used in our experiments. Appendix~\ref{Additional_Experimnetal_analysis} presents additional analyses of the experimental results. Appendix~\ref{appendix:experimental-details} provides detailed experimental settings and implementation details. Appendix~\ref{appendix:hardware-specification} describes the hardware used for the experiments.

\section{Alternative Training Utility Functions}
\label{Appendix:alternative-objectives}

In addition to the training utility functions introduced in ~\eqref{Min-var-utility},~\eqref{Min-sum-sq_advantage-utility}, and~\eqref{Sum-max-R-utility}, we present three alternative training utility functions below.

\paragraph{Minimum of maximum rewards ($U_4$)}
Instead of maximizing the total number of questions with successful trajectories as in~\eqref{Sum-max-R-utility}, we can consider a more conservative objective that focuses on the question with the weakest learning signal in the batch. Specifically, by maximizing the minimum value of the maximum rewards across questions, we encourage the rollout allocation strategy to provide at least one successful trajectory for every question whenever possible. The utility function is therefore defined as
\begin{align}\label{Min-max-R-utility}
    U_4\big(\mathbf{r}(\mathbf{G})\big)
    \triangleq
    \min_{1\le j\le B}
    \max_{1\le i\le G_j} r_j^i .
\end{align}

\paragraph{Summation of reward variances ($U_5$)}
Instead of maximizing the minimum reward variance across questions given in \eqref{Min-var-utility}, we can maximize the total reward variance over the entire batch. Specifically, the training utility function can be defined as
\begin{align}\label{Sum-var-utility}
    U_5\big(\mathbf{r}(\mathbf{G})\big)
    \triangleq
    \sum_{j=1}^{B}\mathrm{Var}_j.
\end{align}
This objective encourages reward diversity across the batch without explicitly balancing the reward variance among questions.

\paragraph{Summation of advantage contributions ($U_6$)}
Instead of maximizing the minimum advantage contribution across questions as in \eqref{Min-sum-sq_advantage-utility}, we can maximize the total advantage contribution of all questions in the batch:
\begin{align}\label{Sum-sum-sq_advantage-utility}
    U_6\big(\mathbf{r}(\mathbf{G})\big)
    \triangleq
    \sum_{j=1}^{B}
    \sum_{i=1}^{G_j}
    \left|A_j^i\right|^2.
\end{align}
Unlike the minimum objective, this utility favors maximizing the overall training signal without explicitly encouraging all questions to contribute equally.

\section{Posterior Update}\label{AppendixA-PosteriorUpdate}

Given the prior belief over $p_j$ as $p_j \sim \mathrm{Beta}(\alpha_j^t,\beta_j^t)$, the probability density function (pdf) of $p_j$ is
\begin{align}
    f(p_j)
    =
    \frac{1}{B(\alpha_j^t,\beta_j^t)}
    p_j^{\alpha_j^t-1}
    (1-p_j)^{\beta_j^t-1},
\end{align}
where the Beta function, $B(\alpha_j^t,\beta_j^t)$, serves as the normalization constant.

Given the observed reward $r_t \in \{0,1\}$, the likelihood follows a Bernoulli distribution:
\begin{align}
    \Pr(r_t \mid p_j)
    =
    p_j^{r_t}
    (1-p_j)^{1-r_t}.
\end{align}
Applying Bayes' rule, the posterior density is proportional to the product of the prior and the likelihood:
\begin{align}
    f(p_j \mid r_t)
    &\propto
    f(p_j)\Pr(r_t \mid p_j) \nonumber\\
    &\propto
    p_j^{\alpha_j^t-1}
    (1-p_j)^{\beta_j^t-1}
    p_j^{r_t}
    (1-p_j)^{1-r_t} \nonumber\\
    &\propto
    p_j^{\alpha_j^t+r_t-1}
    (1-p_j)^{\beta_j^t-r_t}.
\end{align}
The resulting posterior has the kernel of a Beta distribution. Therefore, the pdf of the posterior distribution of $p_j$ is
\begin{align}
    f(p_j \mid r_t)
    =
    \mathrm{Beta}
    \left(
    \alpha_j^t+r_t,\,
    \beta_j^t+1-r_t
    \right).
\end{align}

\section{Information on Tasks and Datasets}\label{Appendix:taskData}

This section presents detailed descriptions of the reasoning tasks examined in this work and the datasets on which each task is evaluated. We also provide an exemplar question from each dataset along with its answer to facilitate understanding of the task in Table \ref{tab:example queries}.

\subsection{Mathematical Reasoning} Mathematical reasoning is widely used to assess the reasoning capabilities of LLMs, as it demands structured multi-step reasoning and precise manipulation of symbolic expressions. We evaluate mathematical reasoning performance using three widely adopted benchmarks: 
\begin{itemize}
    \item GSM8K~\citep{cobbe2021training} is a dataset of multi-step arithmetic word problems that serves as a standard measure for mathematical problem-solving. Its queries cover a range of mathematical topics, including arithmetic calculations, unit conversions, and elementary algebra. 
    \item MATH~\citep{hendrycks2021measuring} is a challenging benchmark of advanced, competition-level math problems ranging from geometry and number theory to probability theory. 
\end{itemize}
For the GSM8K dataset, we use the full training split for training and the official test split for evaluation.
For the MATH dataset, we train the model on the first 4,000 training samples and evaluate it on the MATH500 (500 problems from the MATH test set)~\citep{lightman2024lets}.

\subsection{Logical Reasoning} Logical reasoning is a commonly used task to evaluate an LLM’s deductive reasoning ability over language-based premises. For this task, we use LogiQA~\citep{liu2021logiqa}, a benchmark derived from professional logic examination questions that tests formal reasoning skills such as implication, contradiction, and quantification under complex contextual constraints. For the LogiQA dataset, we train the model on the first 5,000 training samples and evaluate it on the official test split.

\subsection{Commonsense Reasoning} Commonsense reasoning evaluates the model’s ability to apply implicit commonsense knowledge—the general understanding that humans have about the world and how it works, but rarely state explicitly—to reasoning tasks. We assess commonsense reasoning using the CoLoTa dataset~\citep{toroghi2025colota}, which focuses on reasoning over long-tail entities that are underrepresented in LLM pretraining corpora. CoLoTa presents a challenging out-of-distribution (OOD) setting that evaluates the model’s ability to generalize beyond famous entities and contexts. Since CoLoTa does not provide an official train/test split, we randomly partition the dataset, using 90\% of the samples for training and the remaining 10\% as the test set.

\begin{table}[t]
\centering
\small
\setlength{\tabcolsep}{4pt}
\renewcommand{\arraystretch}{1.2}
\caption{Example questions and their answers from the studied evaluation datasets.}
\label{tab:example queries}
\begin{tabularx}{\textwidth}{@{}p{1.6cm} X p{2.4cm}@{}}
\textbf{Dataset} & \textbf{Sample Question} & \textbf{Answer} \\
\midrule
GSM8K &
Five food companies sponsored a local food bank. Foster Farms donated 45 dressed chickens; American Summits donated twice the number of bottled water than the number of dressed chicken donated by Foster Farms; Hormel donated three times the number of dressed chickens that Foster Farms donated; Boudin Butchers donated one-third of the number of dressed chickens that Hormel donated; Del Monte Foods donated 30 fewer bottles of water than American Summits. How many food items did the companies donate in total? &
375 \\

\midrule
MATH &
A parabola has vertex $(4,2)$ and passes through $(2,0).$ Enter the align of the parabola in the form ``$y = ax^2 + bx + c$''. &
$y=-\frac{1}{2}x^2+4x-6$ \\

\midrule
LogiQA & 
\textbf{Article:} The personal right of the author, also known as the moral right of the author, refers to the rights of the copyright owner to his work that are related to the person or are inseparable without direct property content. Reputation, prestige, and the right to maintain the integrity of the work. Copyright property rights, also known as copyright economic rights, are the symmetry of the author's personal rights and refer to the right of the copyright owner to use or authorize others to use the work in a certain way to obtain material benefits. 
\newline \textbf{Question:} According to the above definition, Which of the following statements are correct: 
\newline \textbf{Options:} A. A well-known online novel writer asks the publisher to publish his work with his real name, then the writer exercises his personal rights. B.  The painter Wang sold his painting to Jiang at a high price, and Jiang obtained the copyright of the painting. C. The dance created by a well-known dancer was loved by the masses, and many people imitated and performed commercial performances. The dancer asked to stop this infringement, but she exercised her personal rights. D. A TV drama created by a playwright is remake because it is loved by the audience. The playwright asks the TV show producer not to modify the ending of the story, then he is exercising the property right of copyright. & A \\
\midrule
CoLoTa & Could someone in Vaughan take a taxi to the Jamjuree Art Gallery? & No\\
\bottomrule
\end{tabularx}
\label{tab:dataset_examples}
\end{table}

\section{Additional Experimental Analysis}\label{Additional_Experimnetal_analysis}

In this appendix, we provide additional analyses of the experimental results presented in the main paper.

\subsection{Test Reward During Training}

\subsubsection{Llama-3.1-8B-Instruct model}

\begin{figure}[h]
    \centering
    \begin{subfigure}[t]{0.48\textwidth}
        \centering
        \includegraphics[width=\linewidth]{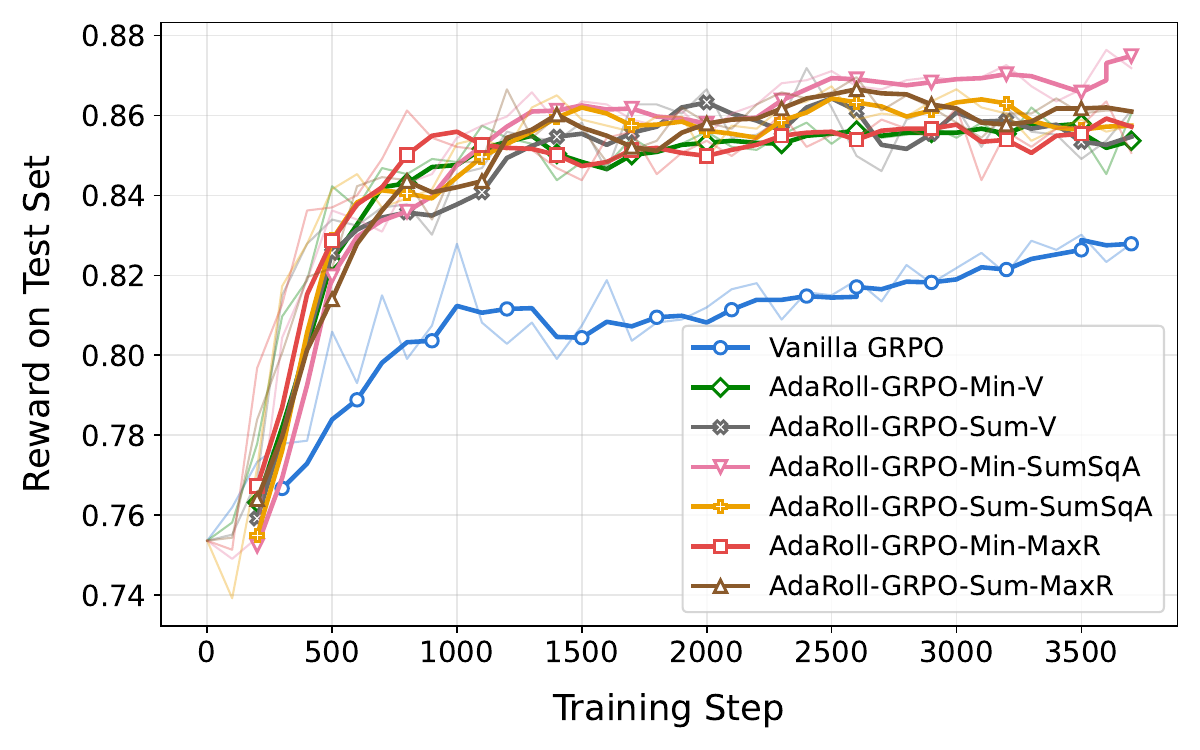}
        \caption{GSM8K dataset.}
    \end{subfigure}
    \begin{subfigure}[t]{0.48\textwidth}
        \centering
        \includegraphics[width=\linewidth]{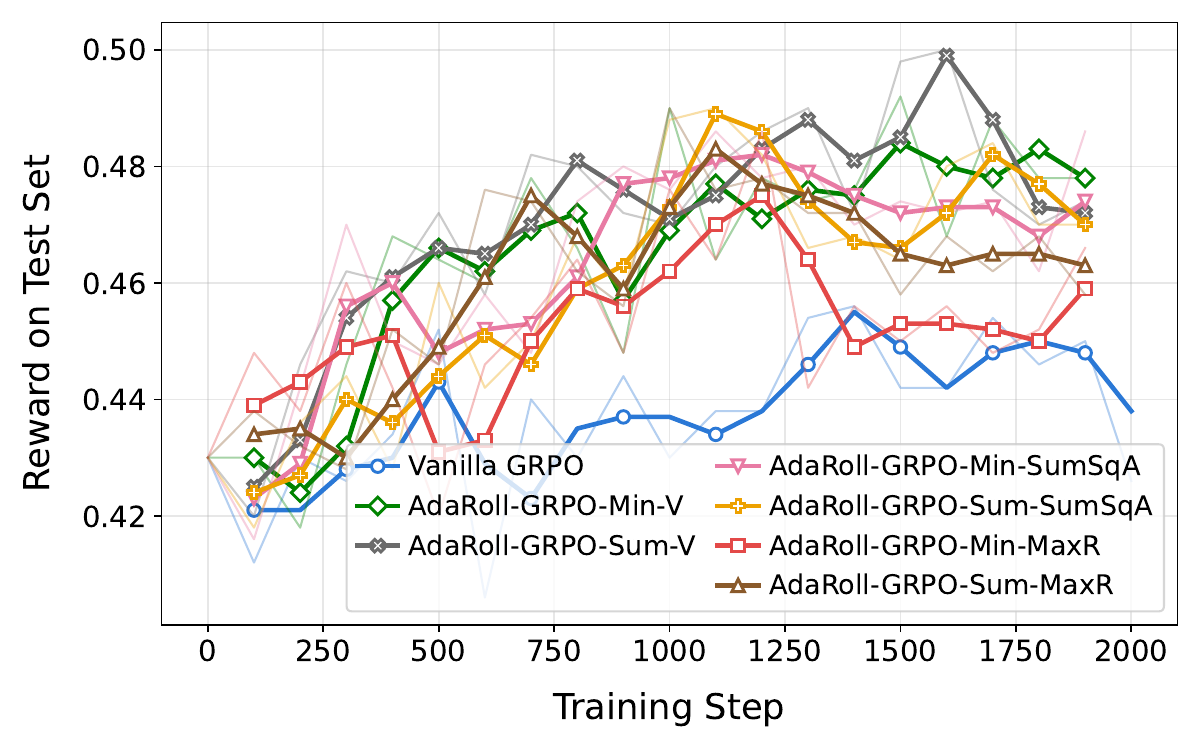}
        \caption{MATH dataset.}
    \end{subfigure}
    \caption{Comparison of the test-set reward achieved by different AdaRoll-GRPO variants and vanilla GRPO throughout training the \textbf{Llama-3.1-8B-Instruct model}. Left: training on the GSM8K dataset. Right: training on the MATH dataset.}
    \label{test_rewards_llama3.1-8b-instrcut}
\end{figure}

To inspect the effect of AdaRoll-GRPO throughout training, we periodically evaluate the models on the test sets. Figure~\ref{test_rewards_llama3.1-8b-instrcut} compares the resulting rewards achieved by different AdaRoll-GRPO variants and vanilla GRPO over the course of training the Llama-3.1-8B-Instruct model. The left and right subfigures report the reward on the GSM8K test set and the MATH500 benchmark, respectively, as training progresses.

As shown in the figure, all AdaRoll-GRPO variants consistently achieve higher test rewards than vanilla GRPO throughout training on both benchmarks, while also attaining higher final rewards. These results demonstrate that adaptively allocating rollouts enables AdaRoll-GRPO to learn more efficiently and effectively, leading to faster performance improvements and better final performance.

\subsubsection{Llama-3.2-3B-Instruct model}

\begin{figure}[h]
    \centering
    \begin{subfigure}[t]{0.48\textwidth}
        \centering
        \includegraphics[width=\linewidth]{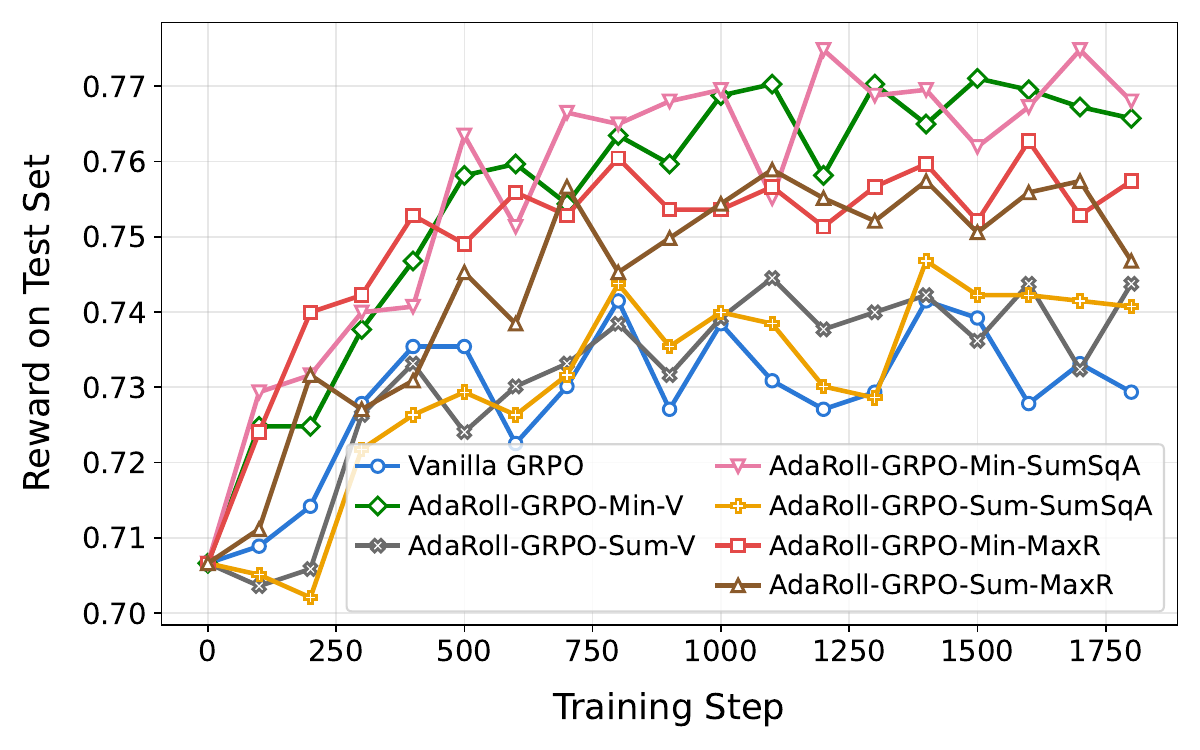}
        \caption{GSM8K dataset.}
    \end{subfigure}
    \begin{subfigure}[t]{0.48\textwidth}
        \centering
        \includegraphics[width=\linewidth]{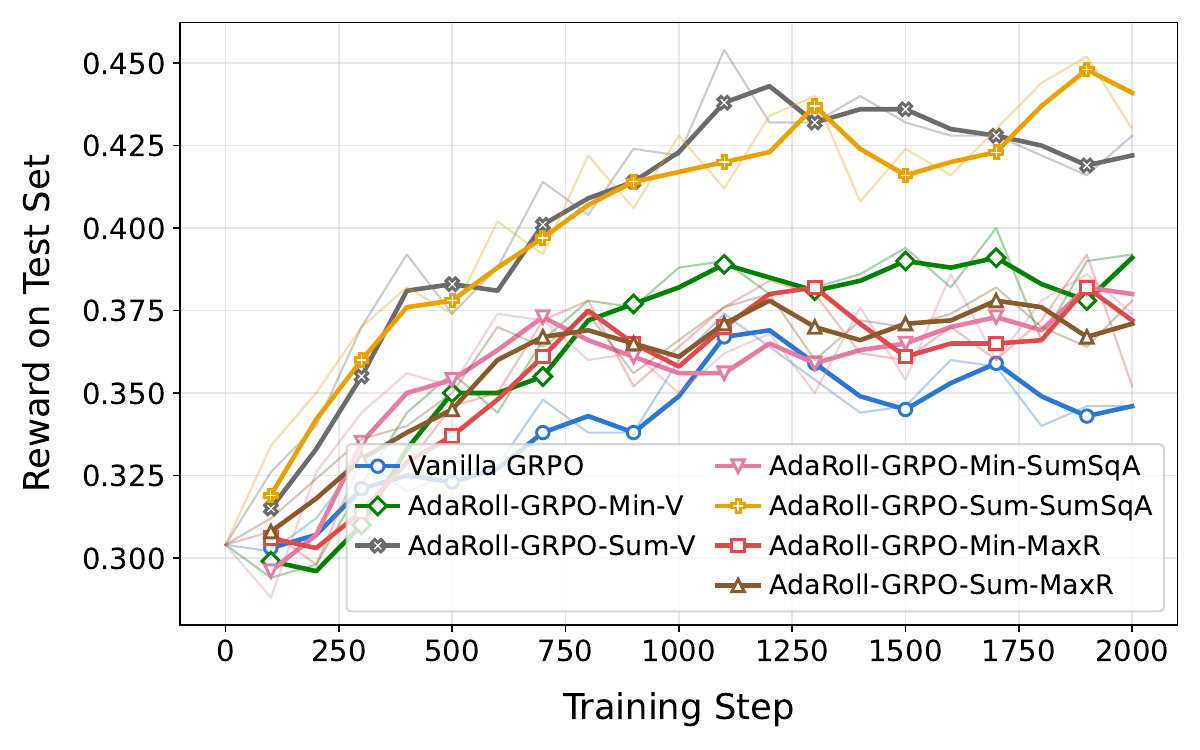}
        \caption{MATH dataset.}
    \end{subfigure}
    \caption{Comparison of the test-set reward achieved by different AdaRoll-GRPO variants and vanilla GRPO throughout training the \textbf{Llama-3.2-3B-Instruct model}. Left: training on the GSM8K dataset. Right: training on the MATH dataset.}
    \label{test_rewards_llama3.2-3b-instrcut}
\end{figure}

Figure~\ref{test_rewards_llama3.2-3b-instrcut} shows the test rewards throughout training for the Llama-3.2-3B-Instruct model. As shown in the left figure, corresponding to the GSM8K dataset, most AdaRoll-GRPO variants, including the minimum-variance training utility used in our experiments, outperform vanilla GRPO and achieve higher test rewards. The right figure corresponds to the MATH dataset, which is more challenging than GSM8K and exhibits greater heterogeneity in question difficulty (as shown in Figure~\ref{Delta_vs_entropy}). On this dataset, all AdaRoll-GRPO variants outperform vanilla GRPO and consistently achieve higher test rewards.

\subsection{Analysis of Learned Rollout Allocation Policies}

\begin{figure}[!h]
\centering
\includegraphics[width=\textwidth]{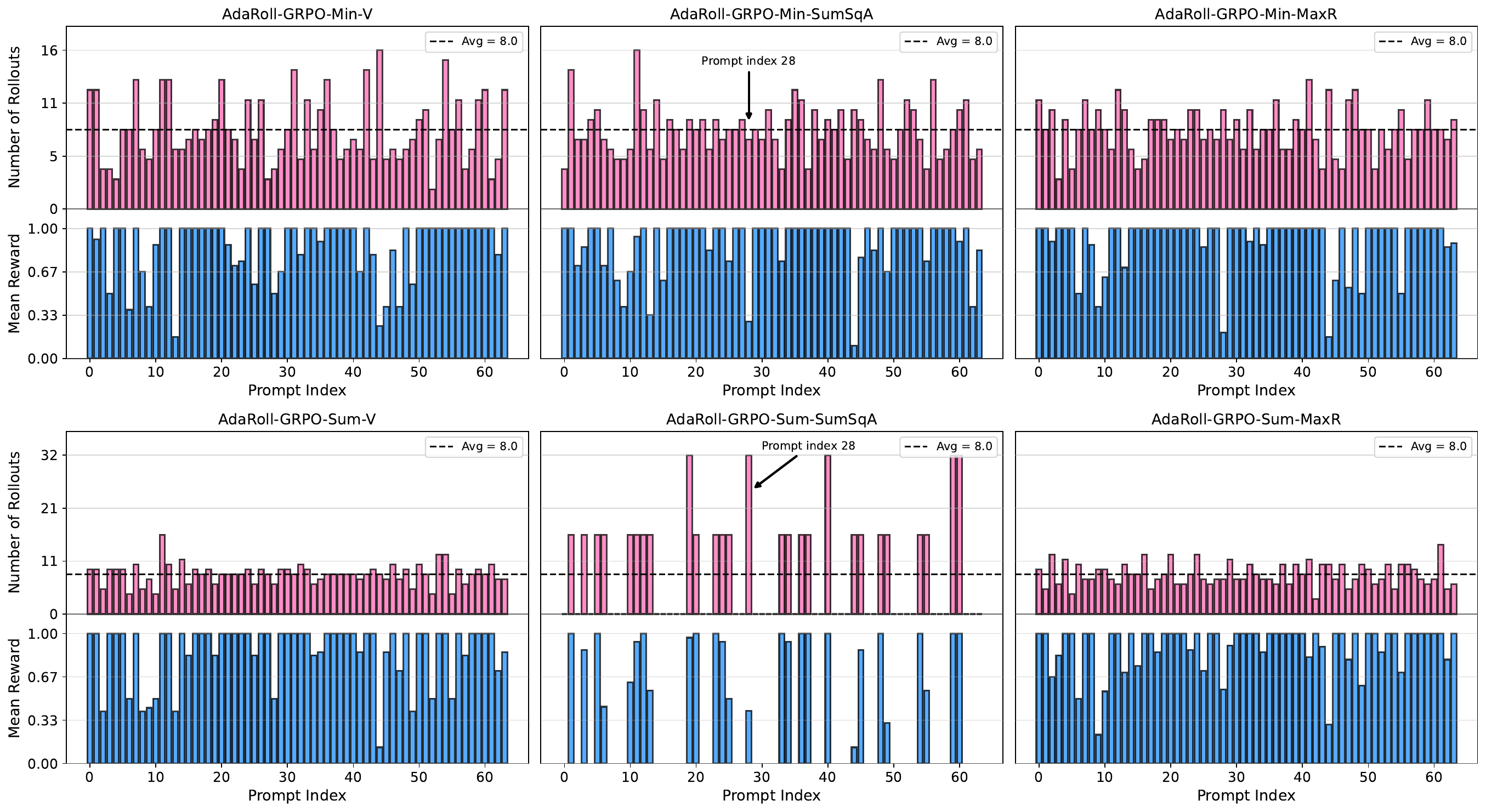} 
\caption{Comparison of rollout allocations across prompts within a batch for different AdaRoll-GRPO variants at a randomly selected training step of \textbf{Llama-3.1-8B-Instruct} on \textbf{GSM8K}, together with the corresponding mean reward of the generated rollouts.}
\label{fig:alloc-gsm8k}
\end{figure}

Figure~\ref{fig:alloc-gsm8k} shows the number of rollouts allocated to
each question in the batch (pink bars, indexed by prompt), together
with the mean reward across that question's rollouts (blue bars),
for six different variants of AdaRoll-GRPO, at a randomly selected training step of Llama-3.1-8B-Instruct on GSM8K.
The horizontal dashed line in each subfigure marks an average of 8 rollouts per question, indicating that the total number of rollouts generated by
AdaRoll-GRPO matches that of vanilla GRPO with $G=8$.


We first compare the results of the utility function defined as the summation of advantage contributions ($U_6$ in \eqref{Sum-sum-sq_advantage-utility}), labeled as \emph{AdaRoll-GRPO-Sum-SumSqA}, with those of the utility function defined as the minimum of advantage contributions ($U_2$ in \eqref{Min-sum-sq_advantage-utility}), labeled as \emph{AdaRoll-GRPO-Min-SumSqA}. Under the \emph{sum} aggregation, many questions receive few or no rollouts, while a small number of questions receive a disproportionately large rollout budget. For example, the question with index 28 receives 32 rollouts under the \emph{sum} utility, whereas it receives fewer than 8 rollouts under the corresponding \emph{min} utility. This question is moderately difficult because its mean rollout reward is slightly above 0.33 in one case and slightly below 0.33 in the other. As a result, it generates both correct and incorrect trajectories, allowing it to contribute substantially to the overall utility. The \emph{sum} aggregation exploits this by overallocating rollouts to such questions. In contrast, the \emph{min} aggregation avoids allocating too many rollouts to any single question, since doing so would reduce the rollout budget available for other questions and could decrease the minimum utility across the batch.

We now compare the results of the utility function defined as the sum of reward variances ($U_5$ in \eqref{Sum-var-utility}), labeled as \emph{AdaRoll-GRPO-Sum-V}, with those of the utility function defined as the minimum reward variance ($U_1$ in \eqref{Min-var-utility}), labeled as \emph{AdaRoll-GRPO-Min-V}. Under the \emph{min} aggregation, many questions with a mean reward of one, corresponding to zero reward variance, receive substantially more than the average rollout budget (equal to 8). Consequently, questions with intermediate mean rewards (between 0 and 1), which have nonzero reward variance, receive fewer rollouts. The \emph{sum}-variance utility exhibits the same qualitative behavior, but much less aggressively. Although questions with mean reward one are still favored, fewer of them are oversampled, and those that are, receive fewer additional rollouts. As a result, the rollout allocation produced by the \emph{sum}-variance utility is more uniform than that of the \emph{min}-variance utility.

We next compare the third utility function, $U_3$ in \eqref{Sum-max-R-utility}, labeled as \emph{AdaRoll-GRPO-Sum-MaxR} and the fourth utility function, $U_4$ in  \eqref{Min-max-R-utility}, labeled as \emph{AdaRoll-GRPO-Min-MaxR}. Both objectives prioritize questions that have not yet produced a successful rollout, since these are the questions suppressing the objective in either aggregation: a question that has already secured a success contributes its maximal reward and can no longer improve $U_3$ or $U_4$, so the learned policies for both variants direct their remaining budget toward the still-unsolved questions in the batch. Under \emph{AdaRoll-GRPO-Sum-MaxR}, the policy concentrates rollouts on the unsolved questions with the highest estimated probability of success, since a single correct rollout on any one of them immediately increases the sum; questions with very low estimated success probabilities are effectively deprioritized, as additional rollouts are unlikely to pay off within the remaining budget. \emph{AdaRoll-GRPO-Min-MaxR} does not explicitly prioritize unsolved questions with the highest success probability, as solving them does not yield an immediate increase in the objective.

\begin{figure}[!h]
\centering
\includegraphics[width=\textwidth]{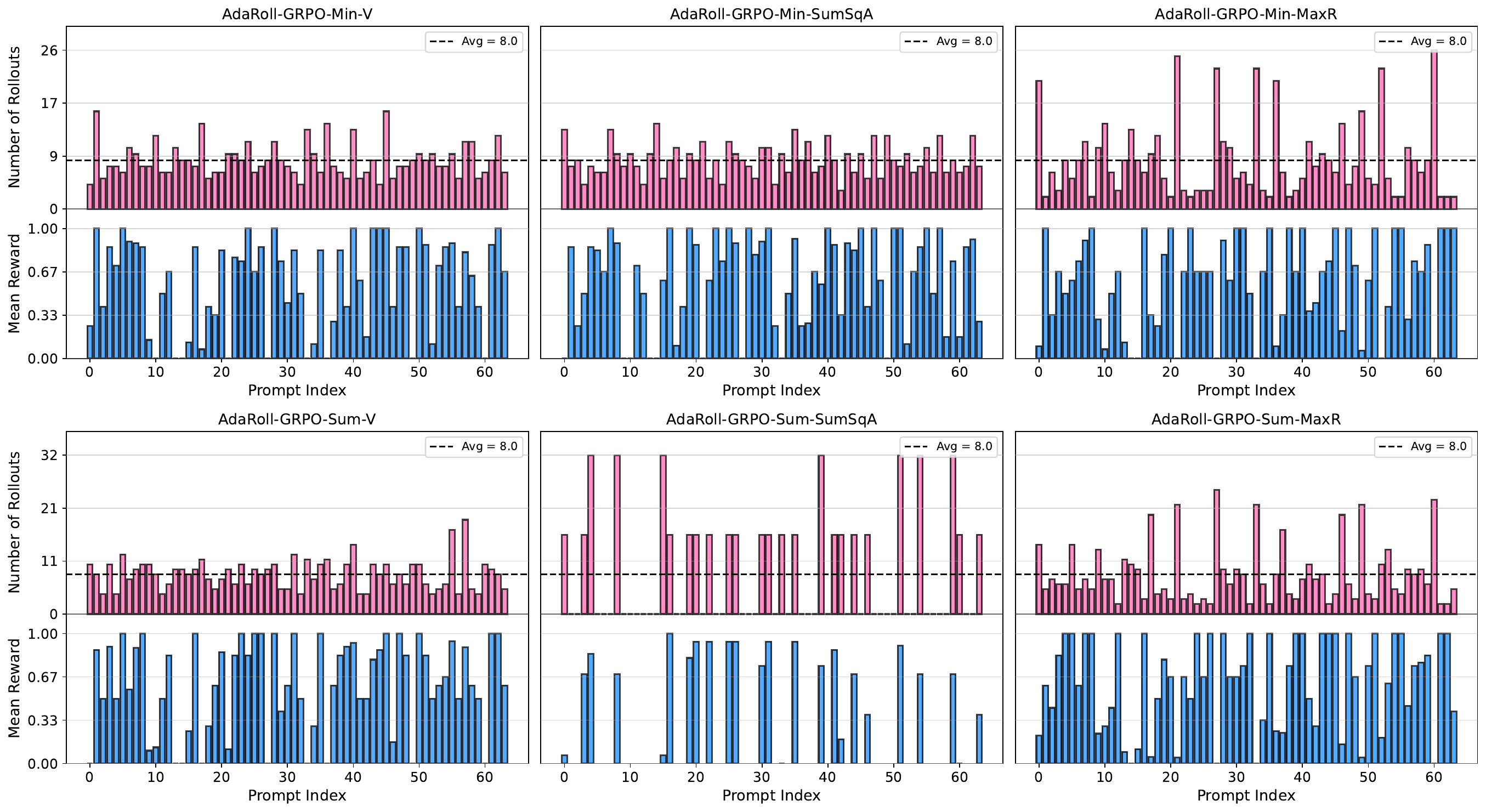} 
\caption{Comparison of rollout allocations across prompts within a batch for different AdaRoll-GRPO variants at a randomly selected training step of \textbf{Llama-3.1-8B-Instruct} on \textbf{MATH}, together with the corresponding mean reward of the generated rollouts.}
\label{fig:alloc-math}
\end{figure}

The same trend is also observed in Figure~\ref{fig:alloc-math}, which shows the rollout allocation policies learned by different AdaRoll-GRPO variants at a randomly selected training step during training of Llama-3.1-8B-Instruct on MATH. 
Under \emph{AdaRoll-GRPO-Sum-SumSqA}, the policy concentrates almost the entire rollout budget on a rather small subset of questions, allocating the maximum of 32 rollouts to several of them while leaving many others with few or no rollouts; \emph{AdaRoll-GRPO-Min-SumSqA} instead spreads the budget far more evenly, with no question receiving more than 15 rollouts. In contrast, \emph{AdaRoll-GRPO-Min-V} allocates disproportionately large budgets to the many questions with mean reward one (zero variance), while \emph{AdaRoll-GRPO-Sum-V} favors these questions less aggressively and distributes rollouts more evenly.

Comparing \emph{AdaRoll-GRPO-Min-MaxR} and \emph{AdaRoll-GRPO-Sum-MaxR}, we observe that \emph{AdaRoll-GRPO-Min-MaxR} allocates rollouts broadly across questions with zero mean rewards, in order to secure at least one success per question and avoid collapsing the batch minimum. \emph{AdaRoll-GRPO-Sum-MaxR}, by contrast, withdraws budget from questions whose observed success rate is too low and instead pushes rollouts toward unsolved questions with higher marginal probability of achieving a first success. We also observe that \emph{AdaRoll-GRPO-Min-MaxR} leaves 7 questions unsolved, whereas \emph{AdaRoll-GRPO-Sum-MaxR} leaves only 4. This difference stems from their optimization objectives. Specifically, \emph{AdaRoll-GRPO-Sum-MaxR} aims to maximize the total number of solved questions, whereas \emph{AdaRoll-GRPO-Min-MaxR} seeks to maximize the minimum of the maximum rewards across questions. Consequently, \emph{AdaRoll-GRPO-Sum-MaxR} adopts a greedier strategy by prioritizing unsolved questions with a higher probability of being solved. In contrast, the \emph{min}-based objective distributes the remaining rollouts more evenly across unsolved questions, since solving only one or a few of them is insufficient to improve the objective. As a result, this strategy tends to leave more questions unsolved by the end.


\section{Experimental Details}
\label{appendix:experimental-details}

\subsection{Rollout Requirement Analysis}
Figure~\ref{MotiveFigure} is generated using the Llama-3.2-3B-Instruct model with \texttt{max\_new\_tokens}$=1024$, \texttt{temperature}$=0.8$, and \texttt{top\_p}$=0.95$. The model is evaluated on the training split of MATH (4{,}000 samples from all subsets) and the full training split of GSM8K.

For each question, we first generate a single rollout and check whether the generated response contains the correct final answer. The fraction of questions solved at this stage corresponds to the bar at $k=1$.
Next, for each question that remains unsolved, we generate one additional rollout and record the questions that are solved for the first time. Their fraction relative to the entire training set corresponds to the bar at $k=2$.

We repeat this procedure for $k=4, 8, 16, 32, 64,$ and $128$. At each stage, only questions that remain unsolved receive additional rollouts until the total number of rollouts reaches $k$. For each value of $k$, the height of the corresponding bar represents the percentage of questions that are solved for the first time after receiving between $\frac{k}{2}+1$ and $k$ rollouts.

\subsection{Training Hyperparameters}

We tune the learning rate for each model on the validation sets of the datasets by searching among $\{ 1\times 10^{-6}, 5\times 10^{-6}, 1 \times 10^{-5}, 2 \times 10^{-5}, 5 \times 10^{-5} \}$ and report the chosen values in Table \ref{tab:hyperparameters}. For the remaining hyperparameters, we use either the default values from the Hugging Face TRL library~\citep{vonwerra2020trl} or the settings recommended by~\citet{yu2026dapo}, as detailed below. We use the AdamW optimizer with a cosine learning rate scheduler. The per-device batch size is set to $2$, with gradient accumulation over $8$ steps. The maximum completion length is $1024$ tokens. Training is performed for one epoch with \texttt{num\_iterations}$=2$, $\beta=0$ (i.e., no KL regularization), and full-parameter fine-tuning. The loss type is set to \texttt{grpo}, $\texttt{dr\_grpo}$, or \texttt{dapo}, depending on the experiment.

In vanilla GRPO, \texttt{num\_generations} corresponds to $G$, the fixed number of rollouts generated for each question in a batch. For AdaRoll-GRPO, \texttt{num\_generations} specifies the \textit{average} number of rollouts per question, while the actual rollout budget assigned to each question is determined adaptively by the learned allocation policy. We set \texttt{num\_generations} to $8$.

\begin{table}[!h]
\centering
\caption{Selected training hyperparameter values.}
\label{tab:hyperparameters}
\setlength{\tabcolsep}{8pt}
\renewcommand{\arraystretch}{1.35}
\begin{tabular}{@{}lr@{}}
\toprule
\textbf{Hyperparameter} & \textbf{Value} \\
\midrule
Optimizer                     & AdamW \\
Scheduler                     & Cosine \\
Fine-tuning                   & Full-parameter \\
Loss type                     & GRPO / DR-GRPO / DAPO \\
\midrule
Learning rate (Llama-3.1-8B-Instruct, Llama-3.2-3B-Instruct) & $1\times10^{-6}$ \\
Learning rate (Gemma3-4B-IT)              & $5\times10^{-6}$ \\
Learning rate (Qwen2-Math-1.5B)            & $2\times10^{-5}$ \\
$\beta$ (KL coefficient)      & $0$ \\
\midrule
Per-device batch size         & $2$ \\
Gradient accumulation steps   & $8$ \\
Maximum completion length     & $1024$ \\
Training epochs               & $1$ \\
\texttt{num\_iterations}      & $2$ \\
\texttt{num\_generations}                   & 4 / 8 / 16 \\
\bottomrule
\end{tabular}
\end{table}

\subsection{Multi-GPU Training}

AdaRoll-GRPO can be readily applied in a multi-GPU setting. 
Following the TRL library implementation, the number of prompts in the global batch across all devices is
\begin{align}\label{eq:num_prompts}
n_p=\frac{B_{\mathrm{dev}}\,A\,D}{G},
\end{align}
where $B_{\mathrm{dev}}$, $A$, and $D$ denote the per-device batch size, the number of gradient accumulation steps, and the number of GPUs, respectively. Here, $G$ denotes the average number of rollouts generated per prompt and is set to the same value as in vanilla GRPO. During rollout generation and gradient computation, each GPU maintains a copy of the entire global batch of prompts.
Accordingly, each GPU constructs a tabular value function with batch size $B=n_p$ and a rollout budget of
$N=n_p\times\frac{G}{D}.$
The tabular value function is constructed only once and then replicated across all GPUs for use during training.

For example, our Llama-3.1-8B-Instruct experiments use $D=2$ GPUs, per-device batch size $B_{\mathrm{dev}}=2$, gradient accumulation steps $A=8$, and average rollout count $G=8$. Substituting these into~\eqref{eq:num_prompts} yields a global batch of $B= n_p = 4$ prompts and a per-GPU rollout budget of $N=16$, for which computing the tabular value function is tractable.

\subsection{Entropy Computation for RQ~\ref{RQ3}}

Figure~\ref{Delta_vs_entropy} (RQ~\ref{RQ3}) shows the relationship between the performance improvement of AdaRoll-GRPO and the entropy of the training dataset for the Llama-3.1-8B-Instruct and Llama-3.2-3B-Instruct models. The performance improvement is defined as the normalized difference between the final accuracy achieved by AdaRoll-GRPO (using the GRPO loss type) and that achieved by vanilla GRPO, as reported in RQ~\ref{RQ2}. The entropy of each dataset for each model is computed from the distribution of rollout requirements before training, obtained using the same procedure illustrated in Figure~\ref{MotiveFigure}.

\section{Hardware Specification}
\label{appendix:hardware-specification}

All model training experiments are conducted using NVIDIA H100 SXM5 GPUs with 80 GB GPU memory each. For training the 8B model, we use two GPUs, but other models are trained using one GPU.

Auxiliary analysis computations are performed using a computer with an Apple M1 chip featuring an 8-core CPU and 16 GB unified memory.




\end{document}